\PassOptionsToPackage{table}{xcolor}
\documentclass[sigconf]{acmart}

\copyrightyear{2026}
\acmYear{2026}
\setcopyright{cc}
\setcctype{by}
\acmConference[MM '26]{Proceedings of the 34th ACM International Conference on Multimedia}{November 10--14, 2026}{Rio de Janeiro, Brazil}
\acmBooktitle{Proceedings of the 34th ACM International Conference on Multimedia (MM '26), November 10--14, 2026, Rio de Janeiro, Brazil}
\acmDOI{10.1145/3767308.3835142}
\acmISBN{979-8-4007-2213-4/2026/11}
\usepackage{graphicx}
\usepackage{amsmath}
\usepackage{booktabs}
\usepackage{multirow}
\usepackage{algorithm}
\usepackage{algpseudocode}
\usepackage{float}
\usepackage{pifont}
\usepackage{enumitem}
\usepackage{array}
\usepackage{xspace}

\usepackage[capitalize]{cleveref}
\crefname{section}{Sec.}{Secs.}
\Crefname{section}{Section}{Sections}
\Crefname{table}{Table}{Tables}
\crefname{table}{Tab.}{Tabs.}

\makeatletter
\def\onedot{\futurelet\@let@token\@onedot}
\def\@onedot{\ifx\@let@token.\else.\null\fi\xspace}
\makeatother

\makeatletter
\AtBeginDocument{%
  \providecommand{\theHALG@line}{\arabic{ALG@line}}%
  \renewcommand{\theHALG@line}{\thealgorithm.\arabic{ALG@line}}%
}
\makeatother

\definecolor{GainGreen}{RGB}{46,125,50}
\definecolor{GainBG}{RGB}{225,242,226}
\definecolor{LossOchre}{RGB}{198,40,40}
\definecolor{LossBG}{RGB}{239,228,194}
\definecolor{AvgNeutral}{RGB}{238,241,245}
\definecolor{TableHeader}{RGB}{227,232,238}
\definecolor{TableStripe}{RGB}{246,248,251}
\definecolor{OurRowBG}{RGB}{231,240,229}
\definecolor{DiscDeltaBlue}{RGB}{0,0,0}
\definecolor{RouteClipBlue}{HTML}{5c6bc0}
\definecolor{RouteQwenPurple}{HTML}{8e24aa}
\definecolor{RouteVerifierGreen}{HTML}{2e7d32}
\newcommand{\gain}[1]{\textcolor{GainGreen}{\textbf{#1}}}
\newcommand{\loss}[1]{\textcolor{LossOchre}{\textbf{#1}}}
\newcommand{\avgbase}[1]{\cellcolor{AvgNeutral}\textbf{#1}}
\newcommand{\avggain}[1]{\cellcolor{GainBG}\textcolor{GainGreen}{\textbf{#1}}}

\newcommand{\discref}[1]{\textcolor{DiscDeltaBlue}{\textbf{#1}}}

\newcommand{\papertitletext}{G2D: Generative-to-Discriminative Collaborative Inference for Zero-Shot Image Classification}
\newcommand{\supplementarylink}{\href{https://github.com/Harzva/G2D/blob/main/paper/G2D_Supplementary.pdf}{supplementary material}}

\begin{document}

\title{\papertitletext}

\author{Zehua Hao}
\affiliation{%
  \institution{Xidian University}
  \city{Xi'an}
  \state{Shaanxi}
  \country{China}}
\email{zhhao\_2025@163.com}

\author{Fang Liu}
\authornote{Fang Liu is the corresponding author.}
\affiliation{%
  \institution{Xidian University}
  \city{Xi'an}
  \state{Shaanxi}
  \country{China}}
\email{f63liu@163.com}

\author{Qinliang Wang}
\affiliation{%
  \institution{Xidian University}
  \city{Xi'an}
  \state{Shaanxi}
  \country{China}}
\email{wangqinliang2025@163.com}

\author{Yaoyang Du}
\affiliation{%
  \institution{Xidian University}
  \city{Xi'an}
  \state{Shaanxi}
  \country{China}}
\email{yaoyangdu@163.com}

\author{Xinyan Huang}
\affiliation{%
  \institution{Xidian University}
  \city{Xi'an}
  \state{Shaanxi}
  \country{China}}
\email{348445373@163.com}

\author{Puhua Chen}
\affiliation{%
  \institution{Xidian University}
  \city{Xi'an}
  \state{Shaanxi}
  \country{China}}
\email{phchen@xidian.edu.cn}
\renewcommand{\shortauthors}{Zehua Hao et al.}

\begin{abstract}
Zero-shot classification needs efficient label retrieval and fine-grained visual reasoning, yet discriminative and generative vision-language models fail in complementary ways.
When CLIP's top-1 prediction is wrong, the correct label often remains in its top-$K$ shortlist, making disambiguation rather than recall the key challenge.
Standalone generative models, however, are hindered by large label spaces and unconstrained outputs.
This complementarity motivates separating broad candidate retrieval from fine-grained, image-grounded verification.
We propose \textbf{G2D}, a training-free framework that uses a generative VLM to verify CLIP-retrieved candidates against the image.
Candidate names and CLIP probabilities provide a structured prior for resolving visually similar classes.
Fixed confidence routing, entropy-adaptive candidate sizing, and trie-constrained decoding focus generative reasoning on uncertain samples and ensure one valid output for each input at test time.
Across eight benchmarks, G2D achieves 68.85\% average accuracy, versus 59.35\% for CLIP and 63.11\% for the standalone VLM.
Across seven generator configurations, candidate-set verification improves average accuracy by 1.08--27.42 percentage points.
G2D also transfers to DCLIP, WaffleCLIP, and CuPL, supporting a practical interface between discriminative proposal and generative visual reasoning.
Code: \url{https://github.com/Harzva/G2D}.
\end{abstract}



\begin{teaserfigure}
\centering
\includegraphics[width=\textwidth]{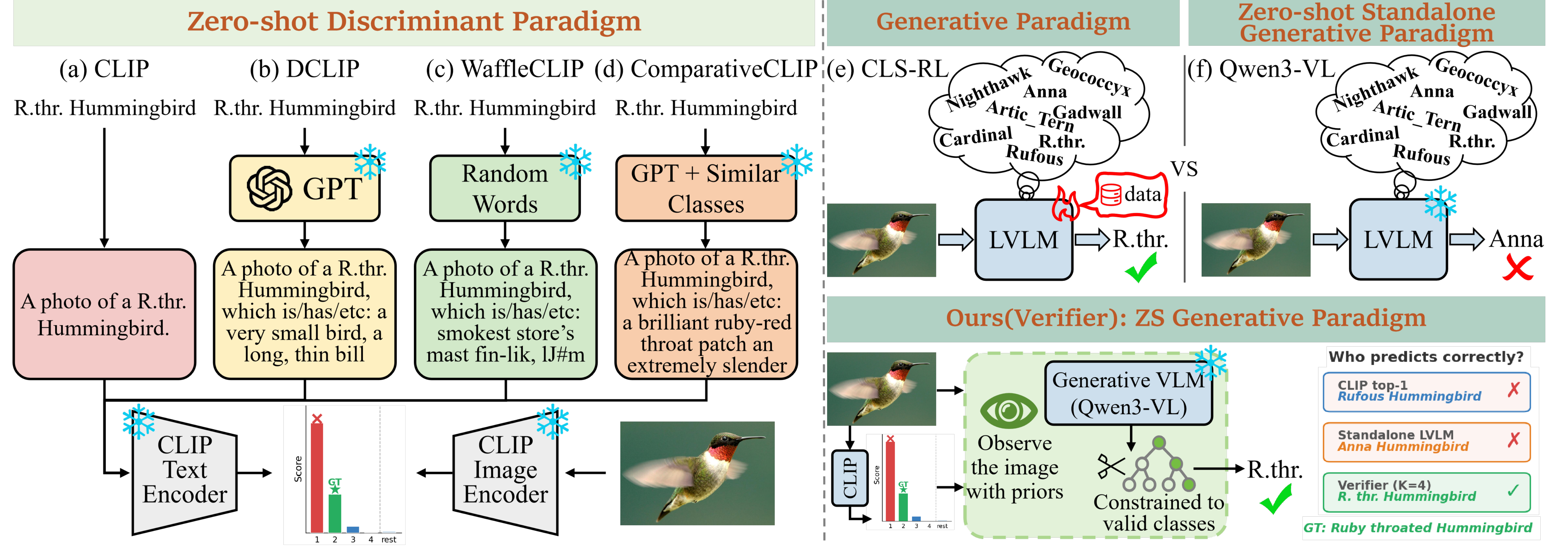}
\caption{Paradigm comparison for zero-shot image classification.}
\Description{A panoramic figure with seven panels. Panels (a)--(d) show discriminant-paradigm pipelines (CLIP, DCLIP, WaffleCLIP, ComparativeCLIP) that rewrite text prompts without seeing the image. Panel (e) shows CLS-RL, a generative paradigm requiring RL training. Panel (f) shows Qwen3-VL applied directly to zero-shot classification, contrasting full-class vs.\ constrained candidate sets. The bottom-right panel shows G2D: a generative VLM constrained to CLIP's top-K, with a CUB200 example where G2D succeeds while CLIP and standalone VLM fail.}
\label{fig:teaser}
\label{fig:motivation}
\end{teaserfigure}

\maketitle

\clearpage
\section{Introduction}
\label{sec:intro}
\begin{figure}[t]
\centering
\includegraphics[width=\columnwidth]{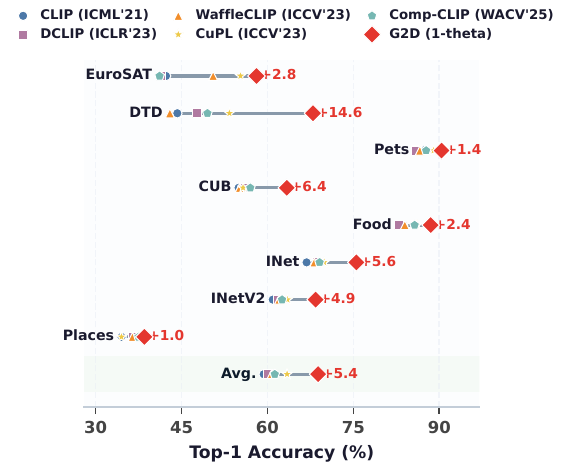}
\caption{\textbf{Prompt-enriched baseline comparison under the shared zero-shot protocol.}
Strict G2D$(1\theta)$ is compared against representative prompt-enriched CLIP baselines (DCLIP~\cite{Menon-2023-DCLIP}, WaffleCLIP~\cite{Roth-2023-Waffle}, CuPL~\cite{Pratt-2023-CuPL}, and Comp-CLIP~\cite{Lee-2025-ComparativeCLIP}) across all eight datasets.}
\Description{A dumbbell plot comparing strict G2D with four prompt-enriched CLIP baselines and vanilla CLIP across eight datasets and their average.}
\label{fig:baseline_compare}
\end{figure}

Zero-shot image classification with CLIP~\cite{Radford-2021-CLIP} is now a widely used baseline, yet it exhibits a recurring failure mode: the ground-truth label frequently appears in CLIP's top-$K$ predictions even when the top-1 is wrong.
This pattern reveals that \emph{candidate coverage is not the bottleneck}; disambiguation under residual uncertainty is.

\textbf{Discriminant paradigm} (\cref{fig:teaser}(a)--(d)).
A growing family of methods enriches CLIP's text prompts with LLM-generated descriptions~\cite{Menon-2023-DCLIP,Pratt-2023-CuPL,Roth-2023-Waffle,Lee-2025-ComparativeCLIP,Wang-2026-RefiningTextPromptZSVAR,Wang-2024-ViLTCLIP,Wang-2025-VLPACLIP,Wang-2026-VCGPrompt,Wu-2025-HVTPG,SunHZP23,LiYTZLLW24}, refining text prototypes but never giving the language model access to the test image.
All gains remain within CLIP's similarity-matching paradigm.

\textbf{Generative paradigm} (\cref{fig:teaser}(e)).
CLS-RL~\cite{Li-2025-CLSRL} applies GRPO reinforcement learning to a generative VLM for classification, demonstrating that generative models can be competitive.
However, this approach requires task-specific training data and RL optimisation, and still relies on soft substring matching for evaluation because unconstrained generation cannot guarantee valid class names.

\textbf{Zero-shot generative investigation} (\cref{fig:teaser}(f)).
Can a generative VLM be applied to zero-shot classification \emph{without any training}?
Our investigation with Qwen3-VL~\cite{Qwen3VL-2025} reveals three key challenges:
\begin{enumerate}[leftmargin=*,topsep=1pt,itemsep=1pt]
\item \textbf{Context-window saturation.} Large label sets (hundreds to thousands of classes) inject tail-class tokens into the VLM's context window, creating distributional noise that degrades classification even on otherwise easy samples.
\item \textbf{Unconstrained output drift.} Over 30\% of unconstrained VLM outputs fall outside the valid label set on fine-grained datasets, making reliable evaluation impossible without ad-hoc soft matching.
\item \textbf{Absence of candidate structure.} The VLM must search from scratch over the entire label space, unable to leverage the discriminative model's ranking prior.
\end{enumerate}
Yet generative VLMs possess a complementary strength that CLIP lacks: they can reason directly from the image with fine-grained visual detail, attending to local patterns that global embeddings miss.
When given a small, high-quality candidate set, this visual reasoning becomes highly effective (\cref{fig:teaser}(f), constrained panel).

\textbf{Why these two families are structurally complementary.}
The three challenges above stem from a fundamental architectural asymmetry.
Discriminative models such as CLIP assign a similarity score to each class independently; the total number of classes does not inflate any input, so adding hundreds of tail classes has negligible effect on top-1 accuracy.
Generative VLMs, by contrast, must represent all candidates inside a single context window that grows with the label set, triggering the saturation and drift problems.
Yet CLIP's ranking errors are \emph{local}: the ground-truth class typically remains within CLIP's top-$K$, even when top-1 is wrong (coverage 77--98\% on the four-dataset routing-analysis subset; see the \supplementarylink).
This asymmetry enables a \emph{mutual gain}: attaching a discriminator as a \emph{structural prior} constrains the generative model to a small candidate set, simultaneously eliminating context noise, enforcing output validity, and providing candidate structure, while the generator corrects CLIP's misranking errors by reasoning directly from the image.
We quantify this with a Complementarity Index (CI) of 56.4\% on average (\cref{sec:complementarity}): when one model fails, the other recovers the correct answer more than half the time;
and the verifier formulation improves every generator included in the matched main-panel comparison (\cref{fig:main_benchmark}(b)), with gains ranging from $+1.1$\,pp for strong standalone models to $+27.4$\,pp for weaker ones.
A complementary four-dataset ambiguity-tail analysis in the \supplementarylink further shows that CLIP's accuracy degrades more steeply than the VLM's in its own low-confidence quintile, supporting confidence-based routing as the outer collaboration strategy.
These observations are also consistent with prior zero-shot recognition work that highlights semantic transfer and structured candidate discrimination beyond top-1 matching~\cite{SongZZHJ20}.

Building on this insight, we propose \textbf{G2D}, a \emph{Discriminative-Generative Collaborative Inference} framework.
The core collaborative branch is an inference formulation in which a generative VLM is used as an \emph{image-grounded verifier} inside a discriminator-defined candidate space, rather than as an open-ended classifier.
Rather than using either model in isolation or only rewriting prompts, G2D assigns complementary roles to the two model families:
\begin{itemize}
\item CLIP provides a \emph{structured candidate space} (top-$K$ classes) and a \emph{calibrated confidence prior} that together define what needs to be verified.
\item The generative VLM performs \emph{constrained visual verification}, attending to the actual image to confirm or overturn CLIP's shortlist rather than searching the full label space from scratch.
\item A fixed confidence router spends the computationally expensive generative budget only on samples where CLIP is uncertain, making strict G2D$(1\theta)$ a two-route collaborative system built around this verifier branch.
\end{itemize}

Under this view, routing, adaptive top-$K$, and prompt priors are not the primary contribution by themselves; they are supporting mechanisms that make the verifier formulation work in practice.
Three design choices distinguish G2D from naive cascading:
(1)~\emph{entropy-adaptive candidate sizing}, which adjusts $K$ to CLIP's uncertainty;
(2)~\emph{CLIP probability injection} into the VLM prompt, giving the generative model access to the discriminative prior;
and (3)~\emph{trie-constrained decoding} that restricts generation to valid class names, eliminating the semantic drift that affects 30\%+ of unconstrained VLM outputs on fine-grained datasets.

This paper focuses on pure zero-shot classification.
We do not use few-shot filtering, validation-set tuning, or training-time adaptation.
The main G2D$(1\theta)$ result uses one fixed hyperparameter configuration shared across all datasets, including a single CLIP prompt template and a pre-specified $\theta_{\mathrm{high}}=0.70$.
Selected two-threshold routing uses benchmark labels and is reported separately as calibrated sensitivity analysis.

\noindent\textbf{Contributions.}
\begin{enumerate}[leftmargin=*,topsep=2pt,itemsep=1pt]
\item G2D, a training-free inference formulation that turns a generative VLM into an image-grounded verifier over a discriminator-defined candidate set.
\item A systematic complementarity study across eight zero-shot benchmarks, showing recoverable error mass, low confusion-pattern overlap, and a strong oracle upper bound between CLIP and Qwen3-VL.
\item A practical collaborative pipeline built from entropy-adaptive $K$, CLIP probability injection, and trie-constrained decoding, with a strict one-threshold deployment protocol that uses no labels for routing selection.
\item Empirical validation across eight benchmarks, representative prompt-enriched CLIP baselines, multiple CLIP variants, and multiple generator backbones, together with explicit separation of strict, fixed-pair, and label-calibrated routing results.
\end{enumerate}

\section{Related Work}
\label{sec:related_brief}
\textbf{Semantic transfer for zero-shot recognition.}
Classical zero-shot learning transfers from seen to unseen categories through attributes, label embeddings, or visual--semantic spaces~\cite{Lampert-2014-Attribute,Akata-2013-Attribute,Frome-2013-Semantic,Akata-2014-Embedding,Socher-2013-Cross,Parades-2015-Embarrass}.
Subsequent work strengthens discriminative semantic representations, graph-based transfer, multi-source semantics, and compositional robustness~\cite{Ye-2017-ZeroShot,YangLXL18,ZhangLT19,SongZZHJ20,Hao-2025-PTSI}, while language descriptions support fine-grained representation learning and part-aware transfer~\cite{Reed-2016-Learning,He-2017-Fine,Elhoseiny-2017-Link}.
CLIP~\cite{Radford-2021-CLIP} changes the practical regime by learning a large-scale image--text space in which arbitrary class names can be scored without task-specific training.

\textbf{Improving discriminative vision--language models.}
Many methods enrich CLIP's class text with language-model-generated descriptions or comparative cues~\cite{Menon-2023-DCLIP,Pratt-2023-CuPL,Roth-2023-Waffle,Lee-2025-ComparativeCLIP,Maniparambil-2023-EnhancingCW,Parashar-2023-Prompting,Saha-2024-Improved,SunHZP23,LiYTZLLW24}.
Related directions inject external knowledge during vision--language learning~\cite{Shen-2022-K-LITE}, introduce parameter-free cross-modal attention~\cite{Guo-2023-CALIP}, or identify likely CLIP errors through prediction consistency and semantic hierarchies~\cite{Ge-2023-ZSGeneralization}.
These approaches strengthen discriminative matching, but they do not use a generative model to inspect each test image and verify a discriminator-proposed shortlist.

\textbf{Generative visual reasoning.}
Generative vision--language pretraining and instruction tuning~\cite{Li-2022-BLIP,Li-2023-BLIP2,Liu-2023-LLaVA,Qwen3VL-2025} enable direct visual question answering and label generation.
For recognition, large multimodal models have been evaluated in zero-shot fine-grained settings~\cite{Atabuzzaman-2025-ZSFG}, captioning has been used to construct vocabulary-free candidates~\cite{Conti-2023-CaSED}, and task-specific reinforcement learning can improve generative classification~\cite{Li-2025-CLSRL}.
However, an open-ended generator does not inherently preserve the legal label space; G2D instead uses the generator as an image-grounded verifier and constrains its output to CLIP's candidates.
More broadly, vision foundation models now support medical image interpretation and segmentation, referring remote-sensing interpretation, and language-guided semantic segmentation~\cite{li2026delving,jiao2026foundation,chai2026recs4r,chai2026like,yan2026language}.

\textbf{Adaptation and collaborative inference.}
Tip-Adapter~\cite{Zhang-2022-Tip-Adapter} and recent prompt- or prototype-based methods~\cite{Li-2025-LDC,Li-2025-PCL,Hao-2026-TextAugVision} exploit labeled support images, while CaFo~\cite{Zhang-2023-CaFo} cascades foundation models in a few-shot setting.
These are adjacent but different protocols.
G2D uses no support images or parameter updates: CLIP supplies candidate structure and a fixed router, and the generative VLM resolves uncertain cases through constrained visual verification.

\section{Method}
\label{sec:method}

\begin{figure*}[t]
\centering
\includegraphics[width=\textwidth]{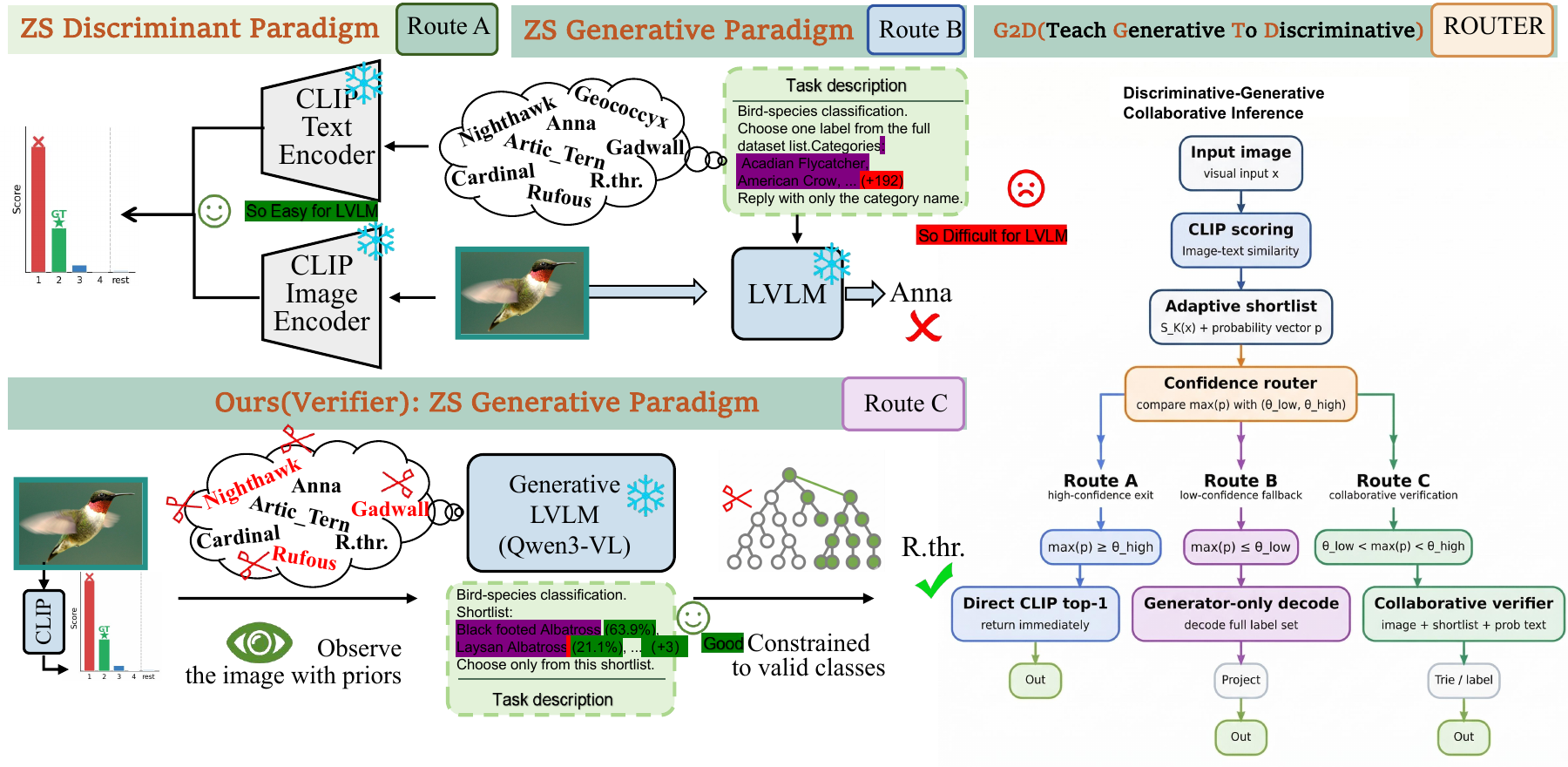}
\caption{\textbf{Unified overview of G2D inference and verifier formulation.}
The figure combines the high-level G2D pipeline with the verifier-centric view.
Pure discriminative zero-shot classification provides useful CLIP priors but can make wrong high-confidence decisions, while pure generative zero-shot classification must reason over the full label set and is more prone to answer drift.
Within G2D, CLIP first constructs an adaptive shortlist and routes the sample by confidence; when verification is invoked, the verifier receives the image together with a CLIP-derived candidate shortlist and prior scores as an input constraint, and its decoding is restricted to valid class names as an output constraint.
This training-free design turns the generative model into a focused visual verifier that corrects errors within CLIP's candidate space.}
\Description{A unified conceptual figure combining the G2D inference framework and the verifier formulation. The figure contrasts discriminative CLIP ranking, standalone generative classification over the full label set, and G2D's routed verifier that uses a CLIP shortlist and prior scores as input constraints together with constrained decoding over valid labels only.}
\label{fig:verifier_task_description}
\end{figure*}

The G2D framework operationalizes the complementarity between discriminative and generative VLMs through three cascaded stages.
We first describe the pipeline (\cref{sec:pipeline}), then detail the three mechanisms that make the collaboration effective (\cref{sec:mechanisms}).
\Cref{fig:verifier_task_description} provides a unified overview: CLIP constructs the candidate prior and routes each sample, while the generative VLM performs image-grounded verification within the constrained candidate space.

\subsection{Problem Setup}
Given an image $x$ and label set $\mathcal{C}$, zero-shot classification predicts
\begin{equation}
\hat{y}=\arg\max_{c\in\mathcal{C}} s\big(\phi_I(x),\phi_T(c)\big),
\end{equation}
where $\phi_I,\phi_T$ are CLIP image and text encoders.
G2D keeps this label space unchanged and only changes inference strategy.

\subsection{G2D Pipeline}
\label{sec:pipeline}

\cref{alg:g2d} summarizes the strict G2D$(1\theta)$ inference procedure used for the main result.

\begin{algorithm}[t]
\caption{Strict G2D$(1\theta)$ Zero-Shot Inference}
\label{alg:g2d}
\begin{algorithmic}
\Require Image $x$, classes $\mathcal{C}$, fixed threshold $\theta_{high}=0.70$
\State Obtain \textcolor{RouteClipBlue}{CLIP} probabilities $\mathbf{p}$ and top-$K(H(\mathbf{p}))$ candidates
\If{$\max(\mathbf{p}) \ge \theta_{high}$}
    \State \Return \textcolor{RouteClipBlue}{CLIP} top-1 prediction \Comment{\textcolor{RouteClipBlue}{Route~A}}
\Else
    \State Invoke collaborative \textcolor{RouteVerifierGreen}{verifier} with candidate-aware instruction
    \State Decode with constrained matching (exact/fuzzy/trie; \cref{alg:trie_decode})
    \State \Return constrained \textcolor{RouteVerifierGreen}{generative} prediction \Comment{\textcolor{RouteVerifierGreen}{Route~B}}
\EndIf
\end{algorithmic}
\end{algorithm}

\subsubsection{Step 1: Discriminative Candidate Extraction (DCE)}
For each image, CLIP provides class probabilities and a candidate set
\begin{equation}
\mathcal{S}_K(x)=\text{Top-}K\big(\mathbf{p}(x)\big)\subset \mathcal{C}.
\end{equation}
This serves a dual role: it \emph{eliminates context noise} for the generative model by reducing the effective label space from $|\mathcal{C}|$ to $K$ (preventing tail classes from inflating the context window and degrading even easy samples) and it \emph{provides a ranked prior} reflecting CLIP's confidence ordering.

\subsubsection{Step 2: Confidence-Aware Dispatch (CAD)}
G2D dispatches each sample based on CLIP top-1 probability.
If the probability exceeds $\theta_{high}$, G2D directly outputs the CLIP prediction (Route~A).
Otherwise, the sample is sent to the collaborative verifier (Route~B).
This fixed two-way rule invokes the expensive VLM only when CLIP confidence is below the pre-specified threshold.
A two-threshold extension adds a generator-only route for very low confidence, but selecting its threshold pairs with benchmark labels makes it calibrated analysis rather than the strict deployment method.
In our experiments, the simple max-probability rule worked better than richer routing signals such as margin or entropy, which did not reliably separate correct from incorrect high-confidence predictions.

\subsubsection{Step 3: Generative Visual Grounding (GVG)}
The verifier route invokes a generative VLM with a CLIP-defined candidate prior.
The verifier uses a \emph{candidate-aware prompt} that includes the candidate class names and, by default, their CLIP probabilities.
The VLM now acts as a verifier: it attends to the actual image and decides among CLIP's candidate classes, while constrained decoding guarantees that the final output stays inside the valid label space (\cref{sec:mechanisms}).
Importantly, this is a direct-label setup: the generator is asked to output the final class label directly, rather than produce an explicit reasoning trace and a separate answer.
We use this no-thinking direct-answer mode by default because enabling implicit thinking changes average accuracy by only $+0.06$ points while increasing wall-clock time per sample by about $1.8\times$, and explicit CoT-think is substantially slower and slightly worse; the matched comparison is documented in the \supplementarylink.

\subsection{Three Mechanisms for Effective Collaboration}
\label{sec:mechanisms}

The naive cascade of ``run CLIP, then run VLM on failures'' is insufficient.
Three mechanisms make the discriminative-generative collaboration effective:

\paragraph{Entropy-Adaptive Candidate Sizing}
Let $\mathbf{p}=(p_1,\dots,p_{|\mathcal{C}|})$ be CLIP class probabilities and $H(\mathbf{p})=-\sum_i p_i\log p_i$ be entropy.
We adapt $K$ by uncertainty:
\begin{equation}
K(H)=
\begin{cases}
K_{\min}, & H\le \tau_{\text{low}}\\
K_{\max}, & H\ge \tau_{\text{high}}\\
K_{\min} + (K_{\max}-K_{\min}) \dfrac{H-\tau_{\text{low}}}{\tau_{\text{high}}-\tau_{\text{low}}}, & \text{otherwise}.
\end{cases}
\end{equation}
When CLIP is moderately confident (low entropy), a small $K$ suffices and the VLM discriminates among fewer candidates.
When CLIP is highly uncertain (high entropy), the candidate set grows to maintain ground-truth coverage.
This adapts the collaborative interface to per-sample difficulty.

\paragraph{CLIP Probability Injection}
The VLM prompt includes both candidate names and their CLIP probabilities.
For example, it may contain ``\texttt{satellite\_dish (23.5\%), television (21.1\%), \ldots}''.
This gives the verifier access to CLIP's distributional prior.
The VLM can override the CLIP ranking when visual evidence contradicts it, but when the prior is informative it helps the verifier avoid ignoring discriminative signals entirely.
Its effect is therefore dataset-dependent: it is most useful when CLIP's confidence is reasonably calibrated, and less useful when that prior is unreliable.

\paragraph{Trie-Constrained Decoding}
Unconstrained VLM generation frequently produces labels outside $\mathcal{S}_K$ or even outside $\mathcal{C}$.
On fine-grained datasets, we observe semantic drift rates of 31.3\% (FGVC-Aircraft) and 30.2\% (Stanford Cars) without constraints.
G2D uses a three-stage label projection: exact match $\to$ fuzzy match $\to$ trie-based prefix-constrained decoding, ensuring the output is always a valid class name in $\mathcal{S}_K(x)$.
This serves a dual purpose: it eliminates semantic drift at inference time, and it enables standard exact-match evaluation identical to discriminative baselines, in contrast to the soft substring matching used by prior generative classifiers~\cite{Li-2025-CLSRL} (see \cref{sec:eval_metric}).
\Cref{alg:trie_decode} gives implementation-level pseudocode, including handling of hyphens and other special symbols; additional projection diagnostics are provided in the \supplementarylink.

\begin{algorithm}[t]
\caption{Trie-Constrained Decoding with Symbol-Aware Normalization}
\label{alg:trie_decode}
\scriptsize
\begin{algorithmic}[1]
\Require Candidate labels $\mathcal{S}_K$, tokenizer $T$, beam size $B$
\State \textsc{Normalize}$(s)$: lowercase; map ``\_''/``-'' to spaces; strip punctuation except $\{/,\&, '\}$; collapse spaces
\State Initialize empty trie $\mathcal{T}$ and map $\mathcal{M}$ (normalized alias $\rightarrow$ canonical label)
\ForAll{$c \in \mathcal{S}_K$}
    \State $\mathcal{A} \gets \{c,\ c[\_ \rightarrow \text{space}],\ c[- \rightarrow \text{space}]\}$
    \State Add $\textsc{Normalize}(a)\mapsto c$ to $\mathcal{M}$ for all $a\in\mathcal{A}$
    \State Insert $T(\text{`` ''}+c)$ and EOS into $\mathcal{T}$ \Comment{keep symbols}
\EndFor
\State Initialize beams with empty token prefix
\For{$t=1,\dots,T_{\max}$}
    \ForAll{beam prefix $\pi$}
        \State allowed $\gets$ valid next tokens from $\mathcal{T}$ under $\pi$
        \If{$\pi$ already matches a complete label in $\mathcal{T}$}
            \State Force EOS; finish this beam
        \ElsIf{allowed is empty}
            \State Map $T^{-1}(\pi)$ by exact/fuzzy match in $\mathcal{M}$; finish
        \Else
            \State Expand beam only with tokens in allowed
        \EndIf
    \EndFor
    \State Keep top-$B$ beams
\EndFor
\State \Return highest-score finished canonical label in $\mathcal{S}_K$
\end{algorithmic}
\end{algorithm}

\subsection{Unified Decision Rule}
Let $y_{\text{clip}}(x)$ be the CLIP top-1 prediction, $y_{\text{gene}}(x)$ be the standalone generator prediction, and $y_{\text{ver}}(x,\mathcal{S}_K)$ be the constrained verifier prediction over $\mathcal{S}_K(x)$.
We define two routing variants.

\paragraph{G2D$(1\theta)$: one-threshold two-way routing.}
A single threshold $\theta_{high}$ splits samples into a CLIP route and a verifier route:
\begin{equation}
\hat{y}(x)=
\begin{cases}
y_{\text{clip}}(x), & \max(\mathbf{p}(x))\ge \theta_{high},\\
y_{\text{ver}}(x,\mathcal{S}_K), & \text{otherwise}.
\end{cases}
\end{equation}
This variant is the simplest instantiation and is used in the discriminator ablation (\cref{tab:disc_ablation_method}) and threshold sensitivity analysis (\cref{fig:disc_theta_sensitivity}) to isolate discriminator-side effects without a generator-only confound.

\paragraph{G2D$(2\theta)$: two-threshold three-way routing (calibrated analysis).}
Two thresholds $(\theta_{high},\theta_{low})$ partition samples among three experts:
\begin{equation}
\hat{y}(x)=
\begin{cases}
y_{\text{clip}}(x), & \max(\mathbf{p}(x))\ge \theta_{high},\\
y_{\text{gene}}(x), & \max(\mathbf{p}(x))\le \theta_{low},\\
y_{\text{ver}}(x,\mathcal{S}_K), & \text{otherwise}.
\end{cases}
\end{equation}
Route~A trusts CLIP when confidence is high; Route~B falls back to the standalone generator when CLIP confidence is very low; Route~C invokes the verifier for intermediate-confidence samples.
G2D$(1\theta)$ is a special case with $\theta_{low}=0$.
The selected G2D$(2\theta)$ pairs use benchmark labels to optimize expert selection and are therefore excluded from strict no-label claims.

In implementation, Route~C decoding is projected to valid labels in $\mathcal{S}_K(x)$ using exact match first, then fuzzy/trie fallback for robust label normalization.
The shared configuration and trie procedure are specified in \cref{sec:exp_setup,alg:trie_decode}.

\subsection{Default Zero-Shot Configuration}
\label{sec:exp_setup}
Main results use the following shared setting across datasets: adaptive $K\in[3,10]$, $\tau_{\text{low}}=0.5$, $\tau_{\text{high}}=2.0$, beam size $=5$, and no validation tuning.
Our primary method is \textbf{G2D$(1\theta)$}, which uses the single shared threshold $\theta_{high}{=}0.70$ across all datasets and routes samples to either CLIP or the verifier.
The threshold is fixed before deployment; no validation or test labels are used to select it.
Unless otherwise noted, \textbf{G2D} in strict zero-shot claims refers to G2D$(1\theta)$.

We additionally report G2D$(2\theta)$ as a routing sensitivity analysis.
Its selected per-dataset threshold pairs are obtained by grid search over cached CLIP-only, generator-only, and verifier-only predictions using benchmark labels; this adds no inference or training, but it is not a no-label deployment protocol.
Exact threshold pairs and protocol-separated diagnostics are provided in the \supplementarylink.

\section{Experiments}
\label{sec:exp}
\subsection{Implementation Details}
\label{sec:impl}
The discriminative backbone is OpenAI CLIP ViT-B/16~\cite{Radford-2021-CLIP} ($\sim$150\,M parameters) with the standard prompt template ``\texttt{a photo of a \{classname\}.}''.
The generative model is Qwen3-VL-8B-Instruct~\cite{Qwen3VL-2025} ($\sim$8.3\,B parameters) loaded in BF16; it operates in direct-label mode (no chain-of-thought) across all evaluations.
We use Qwen3-VL-8B as the default generator to balance accuracy and memory footprint.
All experiments run on 4$\times$ NVIDIA RTX 4090 (24\,GB each).
Verifier outputs are mapped to valid class names via exact match, fuzzy match, and trie-constrained re-generation (\cref{alg:trie_decode}).
All inference is deterministic with a shared core configuration across all datasets; only image resolution and batch size vary.
Further implementation details and complete experimental results are provided in the \supplementarylink; code, configurations, and reproducibility resources are available in our \href{https://github.com/Harzva/G2D}{GitHub repository}.

\subsection{Protocol}
The \emph{main benchmark} evaluates eight datasets: EuroSAT~\cite{Helber-2019-EuroSAT}, DTD~\cite{Cimpoi-2014-DTD}, Oxford-Pets~\cite{Parkhi-2012-Pets}, CUB200~\cite{Peter-2010-CUB200}, Food-101~\cite{Bossard-2014-Food101}, ImageNet~\cite{Deng-2009-ImageNet}, ImageNetV2~\cite{Kornblith-2018-ImageNetv2}, and Places365~\cite{Zhou-2018-Places365}.
For routing diagnostics, visualization, and some historical ablations, we provide deeper analysis on a four-dataset subset (EuroSAT, DTD, Oxford-Pets, CUB200) where matched intermediate artifacts are available.
All experiments are zero-shot.
No few-shot samples are used in the main method.
When prior work reports both zero-shot and few-shot settings, we only consider the zero-shot setting for problem alignment.

\subsubsection{Evaluation Metric}
\label{sec:eval_metric}
Because generative VLMs produce free-form text rather than selecting from a fixed label set, evaluating their classification accuracy requires a label-projection step.
Prior work adopts \emph{soft matching}: CLS-RL~\cite{Li-2025-CLSRL} accepts a prediction as correct if the ground-truth name appears as a substring of the model output (case-insensitive), tolerating extra tokens and formatting variation but potentially over-counting.
G2D takes a stricter approach: rather than relaxing the evaluation metric, we project every generative output onto a valid class name \emph{before} computing accuracy.
The three-stage label projection pipeline (exact match $\to$ fuzzy match with \texttt{SequenceMatcher} $\ge 0.85$ $\to$ trie-constrained re-generation; see \cref{alg:trie_decode} and the \supplementarylink) guarantees that the final prediction is always a member of the candidate set $\mathcal{S}_K(x)$.
Accuracy is then computed by standard exact match against the ground-truth label.
This decouples the evaluation from model-specific formatting quirks and ensures that all reported numbers reflect the same deterministic metric used for CLIP baselines.

\subsubsection{Baselines}
\begin{figure}[t]
\centering
\includegraphics[width=\columnwidth]{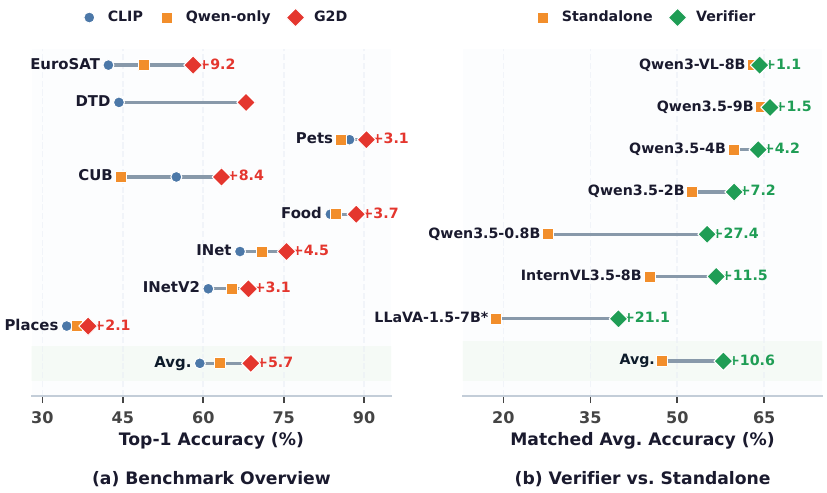}
\caption{\textbf{Benchmark overview and verifier-formulation analysis.}
(a)~CLIP, Qwen-only, and strict G2D$(1\theta)$ are compared across eight datasets and their average.
(b)~The same generators are compared as freeform classifiers (Standalone) and as verifier-only branches over CLIP ViT-B/16 candidates (Verifier~=~G2D w/o routing); LLaVA-1.5-7B is averaged over its six jointly supported datasets.}
\Description{A two-panel figure. Panel (a) compares CLIP, Qwen-only, and strict one-threshold G2D across eight datasets and the average. Panel (b) compares standalone and verifier accuracy for seven generators, with LLaVA averaged on six supported datasets.}
\label{fig:main_benchmark}
\end{figure}

We compare against CLIP top-1 and Qwen-only under the same pure zero-shot protocol.
Descriptor and prompt-enrichment methods are discussed in \cref{sec:related_brief}; the main quantitative table remains focused on methods evaluated under the same pure zero-shot protocol.
In particular, descriptor pipelines that additionally report few-shot filtering variants are treated as external references only in their zero-shot form; few-shot numbers are excluded by design.
Our default CLIP baseline intentionally uses one shared prompt template across datasets rather than prompt ensembles or dataset-specific prompt engineering, so the comparison targets a unified no-tuning zero-shot protocol rather than the strongest prompt-optimized CLIP setting from prior literature.
This choice reflects a deliberate research scope: our goal is not to maximise absolute accuracy by stacking orthogonal engineering tricks, but to measure \emph{how much a collaborative verifier adds on top of a given discriminative--generative pair}.
Because G2D operates as a modular inference wrapper, it produces incremental gains ($\Delta$) over whichever baselines it wraps: \cref{fig:main_benchmark,tab:disc_ablation_method} show this behavior across generator and text-enriched discriminator variants.

\begin{table*}[!t]
\centering
\small
\setlength{\tabcolsep}{3.5pt}
\resizebox{\textwidth}{!}{%
\begin{tabular}{lccccccccc}
\toprule
Method & DTD & CUB200 & Oxford-Pets & EuroSAT & Food101 & Places365 & ImageNetV2 & ImageNet & Avg. \\
\midrule
\textbf{CLIP ViT-B/16-only} & 44.27 & 54.97 & 87.27 & 42.30 & 83.69 & 34.53 & 60.93 & 66.84 & \avgbase{59.35} \\
Qwen3-VL-8B (Gene-only) & 67.97 & 44.70 & 85.75 & 48.93 & 84.82 & 36.38 & 65.32 & 71.01 & \avgbase{63.11} \\
\rowcolor{AvgNeutral}
Verifier-only (no routing) & 52.19 & 59.54 & 90.00 & 47.81 & 87.31 & 37.11 & 66.52 & 73.00 & \avgbase{64.19} \\
\rowcolor{OurRowBG}
\textbf{G2D$(1\theta)$ (strict/no-label)} & \textbf{67.97} & \textbf{63.39} & \textbf{90.41} & \textbf{58.09} & \textbf{88.51} & \textbf{38.52} & \textbf{68.41} & \textbf{75.52} & \avggain{68.85} \\
\midrule
G2D$(2\theta)$ (label-calibrated analysis) & 67.67 & 59.68 & 91.66 & 51.00 & 87.90 & 38.03 & 69.03 & 75.29 & \avgbase{67.53} \\
\bottomrule
\end{tabular}
}
\caption{\textbf{Strict zero-shot result and controlled attribution.}
Top-1 accuracy (\%). The rows are ordered Gene-only $\rightarrow$ Verifier-only $\rightarrow$ fixed routing to isolate full-label generation, candidate-set verification, and routing with the same Qwen3-VL-8B.
G2D$(1\theta)$ fixes $\theta_{\mathrm{high}}=0.70$ before deployment and uses no labeled validation or test feedback.
The selected G2D$(2\theta)$ row uses benchmark labels for threshold selection and is shown only as calibrated analysis.}
\label{tab:main-results}
\end{table*}

\subsection{Main Results}

\Cref{tab:main-results} makes strict G2D$(1\theta)$ the main no-label result.
It reaches 68.85\% average accuracy, improves over CLIP on all eight datasets, and matches or improves Qwen-only on all eight.
With Qwen3-VL-8B fixed, Gene-only (63.11\%) $\rightarrow$ Verifier-only (64.19\%) $\rightarrow$ G2D$(1\theta)$ (68.85\%) isolates full-label generation, candidate-set verification, and fixed routing.
\Cref{fig:baseline_compare} compares G2D with prompt-enriched CLIP methods as a distinct inference paradigm rather than an equal-compute baseline.
The selected G2D$(2\theta)$ row in \cref{tab:main-results} is reported only as label-calibrated sensitivity evidence under the protocol defined in \cref{sec:exp_setup}.

\subsection{Routing, Complementarity, and Mechanism Analysis}
\label{sec:routing_complementarity_mechanism}

\subsubsection{Routing Behavior}
Strict G2D$(1\theta)$ routes high-confidence samples to CLIP and all remaining samples to the verifier.
Across the four-dataset analysis subset, the verifier route retains high candidate coverage and handles 22.9--85.8\% of samples.
\begin{table}[t]
\centering
\small
\setlength{\tabcolsep}{4pt}
\begin{tabular}{lcc}
\toprule
Method & VLM route & Model stack \\
\midrule
CLIP-only & 0\% & $\sim$150M; no VLM \\
Qwen-only & 100\% & Qwen3-VL-8B \\
Verifier-only & 100\% & CLIP + Qwen3-VL-8B \\
\rowcolor{OurRowBG}
G2D$(1\theta)$ & 22.9--85.8\% & CLIP + Qwen3-VL-8B \\
\bottomrule
\end{tabular}
\caption{\textbf{Deployment ledger.} Strict G2D invokes the VLM on 22.9--85.8\% of images on EuroSAT, DTD, Oxford-Pets, and CUB200. This is an accuracy--cost trade-off, not a universal speedup claim.}
\label{tab:cost-ledger}
\end{table}

\subsubsection{Complementarity Analysis}
\label{sec:complementarity}

A key motivation of this paper is that discriminative and generative VLMs make \emph{structurally complementary} errors.
We validate this claim with a systematic analysis spanning all eight zero-shot benchmarks.

\paragraph{Metrics}
Let $A_{\text{clip}}, A_{\text{gen}}$ be the standalone accuracies of CLIP and the generative VLM, and let $A_{\text{oracle}}$ be the accuracy of a per-sample oracle that selects the correct answer whenever \emph{either} model is right.
Let $E_{\text{clip-only}}$, $E_{\text{gen-only}}$, and $E_{\text{both-wrong}}$ denote the fractions of samples on which only CLIP is correct, only the generative VLM is correct, and both models are wrong.
The \emph{Complementarity Index} (CI) measures the fraction of non-shared errors that is recoverable by either model:
\begin{equation}
\text{CI} = \frac{E_{\text{clip-only}} + E_{\text{gen-only}}}{E_{\text{clip-only}} + E_{\text{gen-only}} + E_{\text{both-wrong}}} \times 100\%.
\end{equation}
We also report \emph{confusion-pair overlap}: the overlap between the top-20 confusion pairs induced by CLIP and Qwen. Low overlap indicates diverse failure modes at the error-pattern level.

\paragraph{Results}
The complete complementarity table is provided in the \supplementarylink.
Across eight benchmarks, the oracle reaches 74.62\% average accuracy, compared with 59.35\% for CLIP and 63.11\% for Qwen.
The average CI is 56.4\%, while the top-20 confusion-pair overlap is only 10.3\%, showing substantial recoverable disagreement without implying that routing can attain the oracle.
This pattern is consistent with the matched main-panel comparison in \cref{fig:main_benchmark}(b): the verifier formulation improves every generator included there, with larger gains for weaker standalone models.
A four-dataset ambiguity-tail analysis in the \supplementarylink further shows that the generator's advantage over CLIP peaks in CLIP's low-confidence quintile, supporting the routing strategy.
The remaining gap from the oracle to realized G2D accuracy comes from three practical bottlenecks: CLIP can miss the ground-truth label in top-$K$, routing can leave an ambiguous sample on Route~A, and the verifier can still fail after coverage.
This is why complementarity alone does not guarantee oracle-level performance; the routing diagnostics are needed to interpret the final accuracy--cost trade-off.

\subsubsection{Mechanism Checks}
\label{sec:unified_ablations}
The mechanism-level checks point to a clear split.
Routing mainly controls compute, while CLIP probability text matters most on Oxford-Pets, ImageNet, and Places365.
Both No-Route and No-Prob are verifier-centered analyses rather than generator-only baselines: No-Route sends every sample to the collaborative verifier, while No-Prob removes explicit CLIP probability values from the verifier prompt but retains the CLIP-defined candidate set and constrained decoding.
\subsection{Ablation}
\label{sec:ablation}

\subsubsection{Verifier Formulation vs.\ Standalone Classification}
\label{sec:verifier_ablation}

One key design choice in G2D is to treat the generative VLM as a \emph{constrained verifier} over CLIP's candidate set, rather than as a freeform classifier.
\Cref{fig:main_benchmark}(b) isolates this effect by comparing each generator's standalone accuracy (Gene-only: the VLM classifies from scratch with no CLIP involvement) against the same model used in the verifier-only branch under G2D without outer routing (i.e., every sample is verified).
This comparison removes routing effects and attributes all gains to the verifier formulation itself rather than to expert dispatch.

The gain is positive for every generator included in \cref{fig:main_benchmark}(b), ranging from +1.08 for Qwen3-VL-8B to +27.42 for Qwen3.5-0.8B; LLaVA-1.5-7B gains +21.11 on its six jointly supported datasets.
This suggests that CLIP's candidate structure provides a more valuable scaffold for weaker generators, helping convert unreliable open-label classifiers into more effective verifiers.
Even for the strongest tested generator in this table (Qwen3.5-9B), the verifier formulation still adds a modest positive margin.

\subsubsection{Discriminator Variant}
\label{sec:disc_ablation}

A practical advantage of G2D is its \emph{modularity}: the discriminative backbone can be swapped without retraining or modifying the generative component.
To quantify this, we replace the default CLIP prompt (``\texttt{a photo of a \{class\}}'') with three established text-enrichment strategies: DCLIP~\cite{Menon-2023-DCLIP}, WaffleCLIP~\cite{Roth-2023-Waffle}, and CuPL~\cite{Pratt-2023-CuPL}.
We also evaluate alternative vanilla CLIP backbones (RN50, RN101, ViT-B/32, and ViT-L/14@336) under the same G2D protocol.
Within \cref{tab:disc_ablation_method}, Base, R, and NR use the shared $\theta_{high}{=}0.7$ setting, while 2T is reported only as label-calibrated routing analysis.
The text-enrichment variants are built on ViT-B/16, while the supplementary vanilla-backbone analysis isolates the effect of changing the CLIP visual backbone alone.
\Cref{tab:disc_ablation_method} is kept in the main paper because it directly addresses representative prompt-enriched CLIP baselines; additional vanilla CLIP backbones and one- versus two-threshold comparisons are grouped in the supplementary backbone analysis noted in \cref{sec:impl}.
The table reports standalone discriminator top-1 accuracy, the resulting G2D accuracy, the gain over the corresponding discriminator ($\Delta_{\text{disc}}$), and the gain over the fixed Qwen3-VL-8B-Instruct-only baseline ($\Delta_{\text{gene}}$).
In \cref{tab:disc_ablation_method}, Base denotes Disc-only, R denotes routed G2D$(1\theta)$, NR denotes the verifier-only branch without routing, and 2T denotes the label-calibrated two-threshold variant using per-dataset threshold pairs.
Each tuple $\Delta=(\Delta_{\text{disc}},\Delta_{\text{gene}})$ gives the gains over Disc-only and Qwen3-VL-8B-Instruct-only, respectively.

\begin{table}[t]
\centering
\scriptsize
\setlength{\tabcolsep}{1.6pt}
\resizebox{\columnwidth}{!}{%
\begin{tabular}{@{}llccccc@{}}
\hline
\textbf{Var.} & \textbf{Cfg.} & DTD & CUB200 & Oxford-Pets & EuroSAT & Avg. \\
\hline
\multicolumn{2}{@{}l}{\textbf{Qwen-only}} & \textbf{67.97} & \textbf{44.70} & \textbf{85.75} & \textbf{48.93} & \textbf{61.84} \\
\midrule
\discref{CLIP} & Base & \discref{44.27} & \discref{54.97} & \discref{87.27} & \discref{42.30} & \textbf{57.20} \\
 & R  & 63.24 & 58.68 & 91.66 & 56.14 & \textbf{67.43} \\
\rowcolor{AvgNeutral}
 & $\Delta$  & (\gain{+18.97}, \loss{-4.73}) & (\gain{+3.71}, \gain{+13.98}) & (\gain{+4.39}, \gain{+5.91}) & (\gain{+13.84}, \gain{+7.21}) & (\gain{+10.23}, \gain{+5.59}) \\
 & NR & 64.18 & 58.61 & 92.07 & 56.26 & \textbf{67.78} \\
\rowcolor{AvgNeutral}
 & $\Delta_{NR}$ & (\gain{+19.91}, \loss{-3.79}) & (\gain{+3.64}, \gain{+13.91}) & (\gain{+4.80}, \gain{+6.32}) & (\gain{+13.96}, \gain{+7.33}) & (\gain{+10.58}, \gain{+5.94}) \\
 & 2T & 63.71 & 58.46 & 91.66 & 56.01 & \textbf{67.46} \\
\midrule
\discref{DCLIP} & Base & \discref{47.64} & \discref{55.92} & \discref{86.02} & \discref{41.35} & \textbf{57.73} \\
 & R  & 64.07 & 59.15 & 91.03 & 54.16 & \textbf{67.10} \\
\rowcolor{AvgNeutral}
 & $\Delta$  & (\gain{+16.43}, \loss{-3.90}) & (\gain{+3.23}, \gain{+14.45}) & (\gain{+5.01}, \gain{+5.28}) & (\gain{+12.81}, \gain{+5.23}) & (\gain{+9.37}, \gain{+5.26}) \\
 & NR & 65.54 & 59.20 & 91.20 & 54.26 & \textbf{67.55} \\
\rowcolor{AvgNeutral}
 & $\Delta_{NR}$ & (\gain{+17.90}, \loss{-2.43}) & (\gain{+3.28}, \gain{+14.50}) & (\gain{+5.18}, \gain{+5.45}) & (\gain{+12.91}, \gain{+5.33}) & (\gain{+9.82}, \gain{+5.71}) \\
 & 2T & 64.24 & 58.78 & 91.03 & 54.16 & \textbf{67.05} \\
\midrule
\discref{WaffleCLIP} & Base & \discref{42.97} & \discref{55.16} & \discref{86.62} & \discref{50.52} & \textbf{58.82} \\
 & R  & 64.72 & 59.20 & 91.55 & 56.95 & \textbf{68.11} \\
\rowcolor{AvgNeutral}
 & $\Delta$  & (\gain{+21.75}, \loss{-3.25}) & (\gain{+4.04}, \gain{+14.50}) & (\gain{+4.93}, \gain{+5.80}) & (\gain{+6.43}, \gain{+8.02}) & (\gain{+9.29}, \gain{+6.27}) \\
 & NR & 65.54 & 59.22 & 91.74 & 56.91 & \textbf{68.35} \\
\rowcolor{AvgNeutral}
 & $\Delta_{NR}$ & (\gain{+22.57}, \loss{-2.43}) & (\gain{+4.06}, \gain{+14.52}) & (\gain{+5.12}, \gain{+5.99}) & (\gain{+6.39}, \gain{+7.98}) & (\gain{+9.54}, \gain{+6.51}) \\
 & 2T & 65.31 & 58.89 & 91.55 & 57.05 & \textbf{68.20} \\
\midrule
\discref{CuPL} & Base & \discref{53.01} & \discref{55.75} & \discref{90.30} & \discref{55.30} & \textbf{63.59} \\
 & R  & 66.67 & 58.75 & 92.94 & 57.67 & \textbf{69.01} \\
\rowcolor{AvgNeutral}
 & $\Delta$  & (\gain{+13.66}, \loss{-1.30}) & (\gain{+3.00}, \gain{+14.05}) & (\gain{+2.64}, \gain{+7.19}) & (\gain{+2.37}, \gain{+8.74}) & (\gain{+5.42}, \gain{+7.17}) \\
 & NR & 67.26 & 58.65 & 92.97 & 57.56 & \textbf{69.11} \\
\rowcolor{AvgNeutral}
 & $\Delta_{NR}$ & (\gain{+14.25}, \loss{-0.71}) & (\gain{+2.90}, \gain{+13.95}) & (\gain{+2.67}, \gain{+7.22}) & (\gain{+2.26}, \gain{+8.63}) & (\gain{+5.52}, \gain{+7.27}) \\
 & 2T & 67.02 & 58.27 & 92.94 & 57.47 & \textbf{68.92} \\
\midrule
\discref{\shortstack[l]{Comp.\\CLIP}} & Base & \discref{49.53} & \discref{57.02} & \discref{87.71} & \discref{41.17} & \textbf{58.86} \\
 & R  & 64.89 & 59.58 & 90.68 & 53.64 & \textbf{67.20} \\
\rowcolor{AvgNeutral}
 & $\Delta$  & (\gain{+15.36}, \loss{-3.08}) & (\gain{+2.56}, \gain{+14.88}) & (\gain{+2.97}, \gain{+4.93}) & (\gain{+12.47}, \gain{+4.71}) & (\gain{+8.34}, \gain{+5.36}) \\
 & NR & 66.02 & 59.46 & 90.79 & 53.88 & \textbf{67.54} \\
\rowcolor{AvgNeutral}
 & $\Delta_{NR}$ & (\gain{+16.49}, \loss{-1.95}) & (\gain{+2.44}, \gain{+14.76}) & (\gain{+3.08}, \gain{+5.04}) & (\gain{+12.71}, \gain{+4.95}) & (\gain{+8.68}, \gain{+5.70}) \\
 & 2T & 65.07 & 59.49 & 90.68 & 53.62 & \textbf{67.22} \\
\hline
\end{tabular}
}
\caption{\textbf{Discriminator ablation with text-enrichment methods.}
Four-dataset accuracy (\%) with Qwen3-VL-8B-Instruct, fixed $\theta_{high}{=}0.7$, and adaptive $K{\in}[3,10]$.}
\label{tab:disc_ablation_method}
\end{table}

Two trends emerge from \cref{tab:disc_ablation_method}.

\paragraph{Both Routed and No-Route Variants Remain Beneficial}
On the four evaluated datasets, every recorded $\Delta$ for vanilla CLIP is positive.
The gains are notably large on DTD (+18.97 routed / +19.91 w/o routing) and EuroSAT (+13.84 / +13.96), while Oxford-Pets and CUB200 still improve despite stronger standalone discriminators.

\paragraph{The Discriminator Axis Remains Modular}
The same wrapper remains effective after swapping the discriminator prior from vanilla CLIP to DCLIP, WaffleCLIP, CuPL, and Comparative-CLIP.
Across the four-dataset table, both routed and no-route variants stay positive.
WaffleCLIP yields the largest DTD gain (+21.75 routed / +22.57 w/o routing), Comparative-CLIP provides another useful CUB-oriented prior with +14.76 w/o-routing gain over the generator on CUB200, and CuPL provides the strongest standalone prior and therefore smaller but still consistent collaborative gains.

\begin{figure}[t]
\centering
\includegraphics[width=\columnwidth]{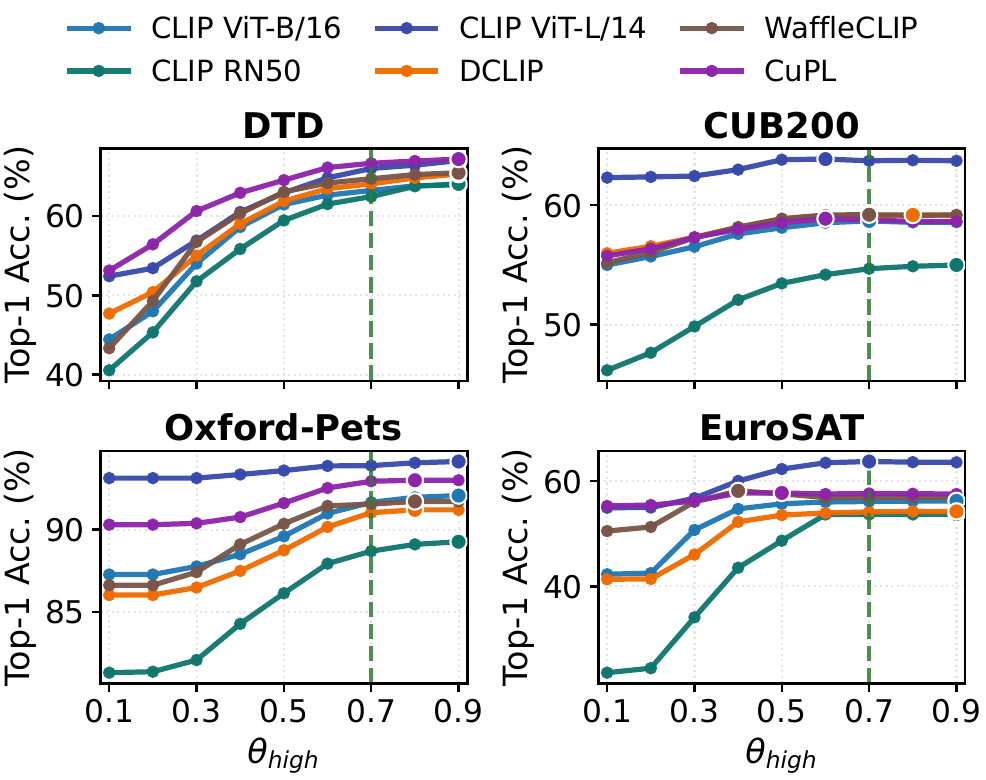}
\caption{\textbf{$\theta_{high}$ sensitivity across discriminator variants with Qwen3-VL-8B fixed.}
The green dashed line marks the shared default $\theta_{high}{=}0.7$; curves are smooth and their optima mostly lie within 0.7--0.9.}
\Description{A threshold-sensitivity plot over discriminator variants and four datasets. The default threshold at 0.7 is marked by a green dashed line, the curves are smooth, and the best points cluster near 0.7 to 0.9.}
\label{fig:disc_theta_sensitivity}
\end{figure}

\paragraph{Discriminator-Side $\theta_{high}$ Sensitivity Is Stable}
To check whether the G2D$(1\theta)$ gains in \cref{tab:disc_ablation_method} are overly dependent on a single routing threshold, we sweep $\theta_{high}\in\{0.1,\dots,0.9\}$ for all six discriminator variants using discriminator priors and the matched no-routing predictions.
For a fair main comparison, all reported routed results retain the shared $\theta_{high}{=}0.7$ rather than selecting a threshold per method or dataset.
\Cref{fig:disc_theta_sensitivity} shows that the curves are generally smooth rather than brittle, and the optimal thresholds cluster in a narrow high-confidence regime for most methods.
Quantitatively, moving from the shared default $\theta_{high}{=}0.7$ to per-curve best thresholds improves performance by only $0.40$ points on average across the 24 method--dataset combinations, with a median gain of $0.17$ and a maximum gain of $1.60$ on CLIP RN50 for DTD.

\subsubsection{Generator Backbone}
\label{sec:gen_ablation}

We next examine whether the collaborative formulation transfers across generator backbones and model scales.
Complete Gene-only, verifier-only, and routing comparisons are included in the grouped supplementary backbone analysis.
They show where candidate-set verification transfers and where weaker or mismatched VLMs remain a limitation; two-threshold rows are treated as routing analysis rather than strict no-label evidence.

Representative examples of the main routing outcomes and the separate label-calibrated G2D$(2\theta)$ analysis are provided in the \supplementarylink.
	
\subsection{Limitations}
\label{sec:discussion}
G2D requires both CLIP and a generative VLM at inference time, and generative inference dominates latency when the verifier-route ratio is high.
This is a deliberate accuracy--cost trade-off: strict G2D$(1\theta)$ eliminates task-specific training and per-dataset routing selection, but it is not computationally equivalent to CLIP-only.
The method also assumes closed-vocabulary recognition and a useful discriminative proposal prior; if the ground-truth class is absent from the shortlist or the prior is misleading, verification can fail.
DTD is a boundary case because CLIP probability text is poorly calibrated for texture recognition, while EuroSAT illustrates a different trade-off: strict G2D improves over Qwen-only (58.09\% versus 48.93\%) but still incurs VLM cost.
The corresponding diagnosis is provided in the \supplementarylink.

\section{Conclusion}
\label{sec:conclusion}
We introduced G2D, a training-free framework that uses a generative VLM as an image-grounded verifier over a discriminator-defined shortlist.
Entropy-adaptive candidate sizing, fixed confidence routing, prior injection, and trie-constrained decoding exploit complementary errors while preserving legal labels.

With $\theta_{\mathrm{high}}=0.70$ fixed before deployment, strict G2D$(1\theta)$ uses no labeled validation or test feedback and reaches 68.85\% average top-1 accuracy across eight benchmarks, improving over CLIP and matching or exceeding Qwen3-VL throughout.
Controlled comparisons attribute the gains to verification and routing.

These results show complementary roles: CLIP supplies efficient shortlist recall, while the verifier resolves residual ambiguity through image-grounded reasoning.
This division of labor combines discriminative retrieval and constrained generation without task-specific training, while preserving valid predictions.

\clearpage

\begin{acks}
This work was supported in part by the National Science and Technology Major Project of the Ministry of Science and Technology of China (Nos.~2025ZD0551500 and 2025ZD0551502); the Key Project of the National Natural Science Foundation of China (Nos.~62431020 and 62231027); the National Natural Science Foundation of China (No.~62576264); the Joint Fund Project of the National Natural Science Foundation of China (No.~U22B2054); the Fund for Foreign Scholars in University Research and Teaching Programs (the 111 Project) (No.~B07048); the Postdoctoral Fellowship Program of China Postdoctoral Science Foundation (CPSF) (No.~GZC20232033); the Program for Cheung Kong Scholars and Innovative Research Team in University (No.~IRT 15R53); and the Key Scientific Technological Innovation Research Project by the Ministry of Education and the National Key Laboratory of Human-Machine Hybrid Augmented Intelligence, Xi'an Jiaotong University (Nos.~HMHAI-202404 and HMHAI-202405).
\end{acks}

\bibliographystyle{ACM-Reference-Format}
\balance
\bibliography{egbib}

\end{document}